\documentclass{article}
\usepackage{spconf,amsmath,graphicx}

\usepackage{subfigure}
\usepackage{amsmath}
\usepackage{amssymb}
\usepackage{hyperref}
\usepackage{booktabs}

\usepackage{multirow}
\usepackage{enumitem}
\usepackage{bbding}
\usepackage{amsfonts}
\usepackage{amsthm}
\usepackage{adjustbox}

\title{Domain-adaptive and Subgroup-specific Cascaded Temperature Regression for Out-of-distribution Calibration}
\name{Jiexin Wang, Jiahao Chen, Bing Su\sthanks{Corresponding author.}}
\address{Gaoling School of Artificial Intelligence, Renmin University of China\\
Beijing Key Laboratory of Big Data Management and Analysis Methods, Beijing, China}
\begin{document}
%
\maketitle
\begin{abstract}
Although deep neural networks yield high classification accuracy given sufficient training data, their predictions are typically overconfident or under-confident, i.e., the prediction confidences cannot truly reflect the accuracy. Post-hoc calibration tackles this problem by calibrating the prediction confidences without re-training the classification model. 
However, current approaches assume congruence between test and validation data distributions, limiting their applicability to out-of-distribution scenarios. 
To this end, we propose a novel meta-set-based cascaded temperature regression method for  
post-hoc calibration. Our method tailors fine-grained scaling functions to distinct test sets by simulating various domain shifts through data augmentation on the validation set. We partition each meta-set into subgroups based on predicted category and confidence level, capturing diverse uncertainties. A regression network is then trained to derive category-specific and confidence-level-specific scaling, achieving calibration across meta-sets. 
Extensive experimental results on MNIST, CIFAR-10, and TinyImageNet demonstrate the effectiveness of the proposed method.
\end{abstract}

  \section{Introduction}
Deep neural networks trained from massive labeled samples can achieve excellent classification performance, but the reliability of the predictions should also be taken into account when these models are applied~\cite{nguyen2015deep,guo2017calibration}. For instance, in risk-sensitive scenarios such as autonomous driving and medical diagnosis, wrong predictions may lead to a crash with catastrophic consequences. 
Therefore, apart from accuracy, accurate prediction confidences are also desirable so that some uncertain predictions can be rejected. 
Besides, in real-world industries, the application scenarios are dynamically changing, and the intelligent system should have access to compute a reliable estimate of the predictive uncertainty in different domains. This requires that the confidence scores of the model should be well calibrated not only for in-distribution (ID) predictions but also for out-of-distribution (OOD) predictions.

\begin{figure}[t]
  \centering
    \subfigure[Most existing methods.]{\includegraphics[width=0.45\linewidth]{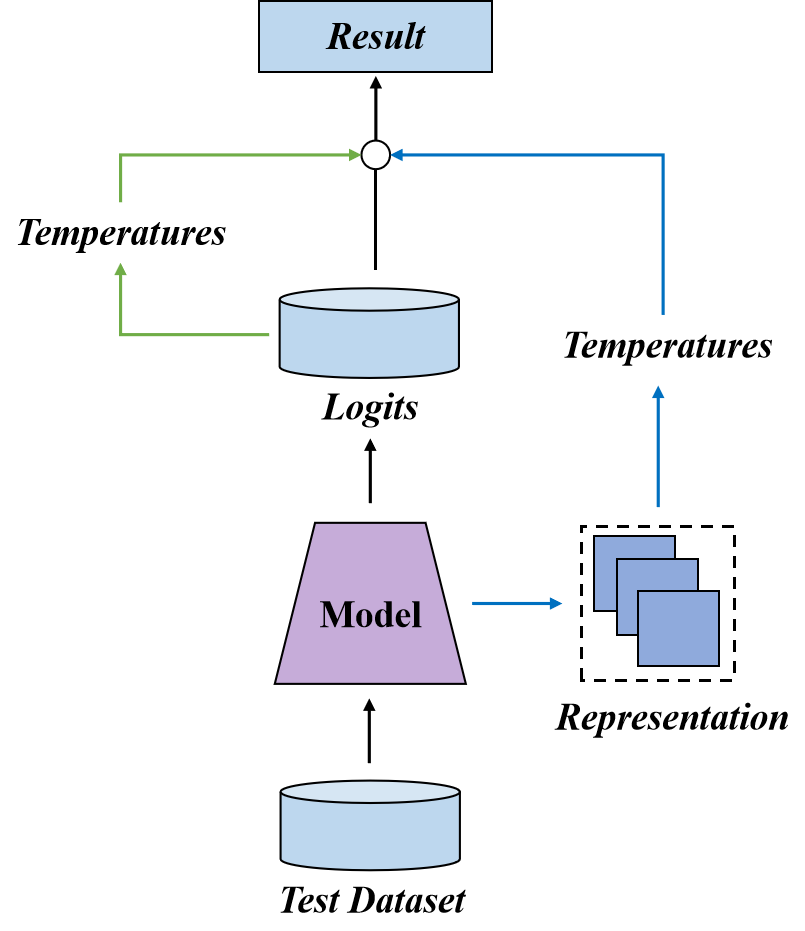}\label{Most existing methods}}
   \subfigure[Our method.]{\includegraphics[width=0.54\linewidth]
   {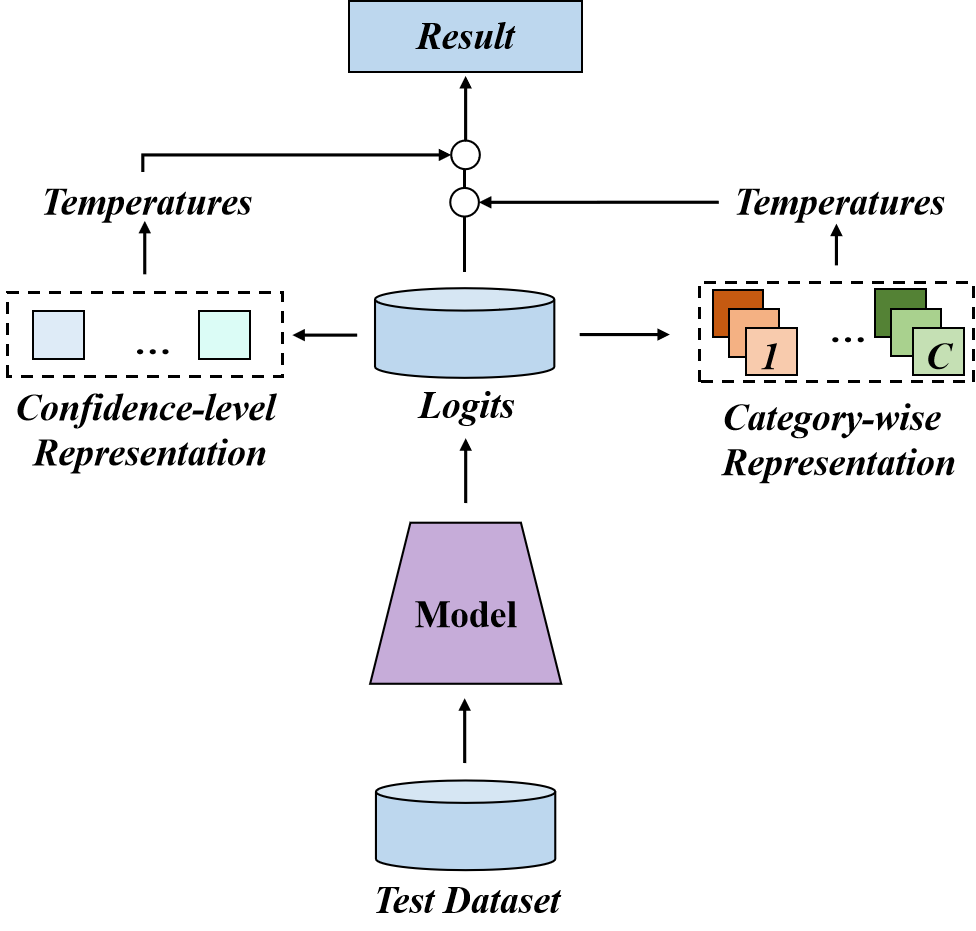}\label{Our method.}}
   \vspace{-0.2in}
   \caption{Comparison of calibration methods. (a): Most existing methods only re-scale the confidence scores using the instance features (blue lines) or the logits (green lines). 
   (b): Our cascaded calibration successively re-scale the logits based on the statistics of instance confidence scores from subgroups of different predicted categories and confidence levels. }
   \label{fig:method_compare}
   \vspace{-0.2in}
\end{figure}

It has been shown that neural network-based classification models tend to yield over-confident predictions and many post-hoc calibration methods, where a validation set drawn from the generative distribution of the training data is used to rescale the outputs of a trained neural network, have been proposed to tackle this problem, including parametric methods~\cite{platt1999probabilistic,niculescu2005predicting,zadrozny2002transforming,guo2017calibration,zhang2020mix} (as shown in Fig.~\ref{Most existing methods}) and non-parametric approaches~\cite{zadrozny2001obtaining,naeini2015obtaining,wenger2020non}. 
Nonetheless, these methods assume that the test data adhere to the same distribution as the training and validation sets, and thus apply a single trained re-scale function to different test sets. However, for OOD scenarios, test sets with different distributions should be calibrated with different re-scale functions since the prediction accuracy and uncertainty of the model are different.

\begin{figure}[t]
    \centering
   \includegraphics[width=0.35\linewidth]{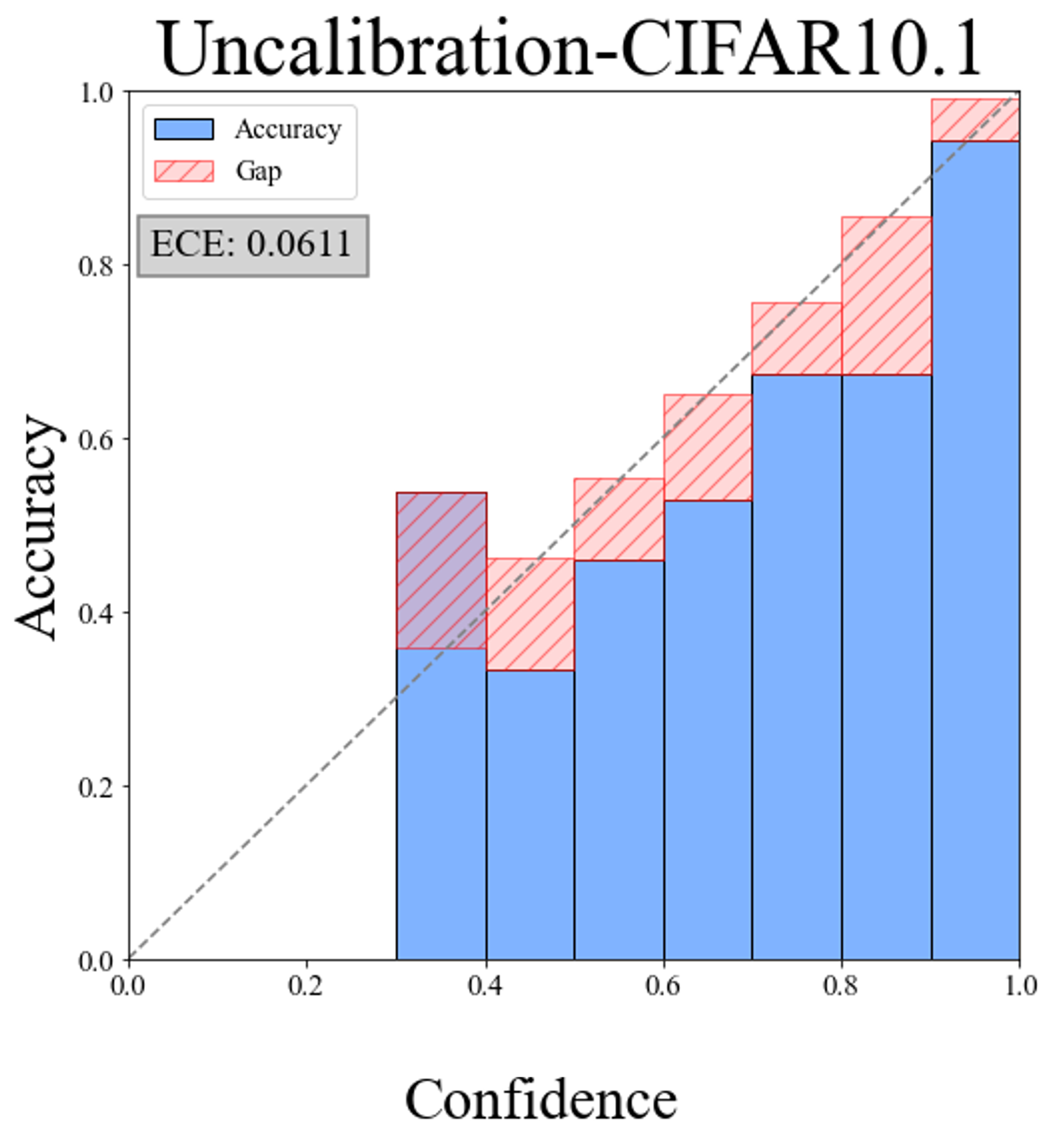} \hspace{5mm}
   \includegraphics[width=0.35\linewidth]{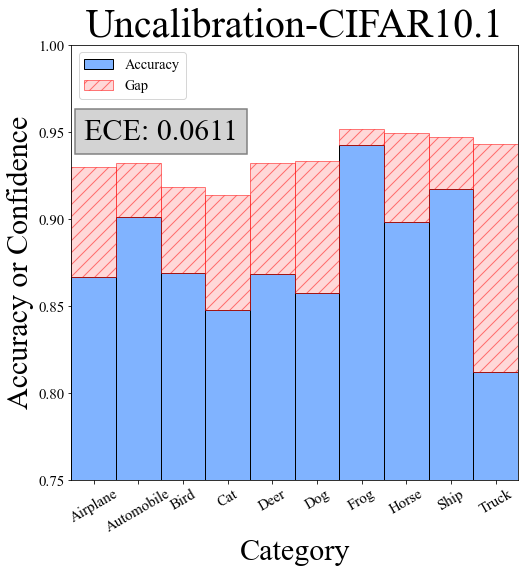}\\
   \vspace{-0.15in}
    \caption{
    Reliability Diagrams on CIFAR10.1: (left) the gap between accuracy and average confidence under different confidence levels and (right) the gap under different predicted categories. ECE is the expected calibration error~\cite{naeini2015obtaining}.}
    \label{fig:uncal_ece_cifar101}
    \vspace{-0.15in}
\end{figure}

There also exist some calibration methods~\cite{liang2017enhancing,jiang2018trust,papernot2018deep,tomani2021post,yu2022robust,liu2022devil} to enhance the reliability of OOD predictions by employing the observations or adding small perturbations in the input. However, such methods apply the same re-scaling function to all data in a test set. Not only do test data with different distributions require different re-scaling functions, but we also observe that the deviation between accuracy and uncertainty is different for instances with different predicted categories and confidence levels. 
Illustrated in Fig.~\ref{fig:uncal_ece_cifar101}, the gaps between accuracy and confidence levels or predicted categories exhibit inconsistency. 
Consequently, the calibration of all predictions through a single scaling function becomes arduous.
Distinct scaling functions are necessary for predictions falling within diverse confidence levels and categories.

In this paper, we introduce a cascaded temperature regression method. Our method regresses specific temperatures from subgroups of the predicted confidences on each test set. Specifically, we generate distinct meta-sets by augmenting the validation set, extracting global representations, and using a regression neural network for temperature regression. The regression network is trained by encouraging calibrated predictions on all meta-sets. In this way, as shown in Fig.~\ref{Our method.}, our method can perform adaptive re-scaling for data in different distributions.
Considering varied impacts of distinct predicted categories and confidence levels on reliability, 
as depicted in Fig.~\ref{fig:uncal_ece_cifar101}, 
we split instances into different subgroups for category-wise and confidence-level-wise calibration. For the former, instances with the highest confidence scores for the same category form subgroups. Confidence score statistics within each subgroup serve as features inputted to the regression network for subgroup-specific temperature regression. This facilitates distinct temperature scaling for predictions across subgroups.
Similarly, a subgroup-specific calibration is performed for different confidence levels.  
The contributions of this paper are summarized as follows:
\begin{enumerate}
    \item We propose a meta-set-based temperature regression method to achieve domain-adaptive re-scaling of the confidence scores, which tackles the OOD post-hoc calibration problem.
    \item We propose a cascaded calibration mechanism to perform subgroup-specific re-scaling by encoding predictions with different predicted categories and different confidence levels into hierarchical representations.
    \item We conduct extensive experiments on different datasets using various architectures.
    The empirical evaluations demonstrate that the proposed method is effective and robust for calibration under OOD scenarios.
\end{enumerate}

 \section{Method}
\subsection{Problem definition}\label{Problem definition}
\begin{figure}[t]
  \centering
   \includegraphics[width=1\linewidth]{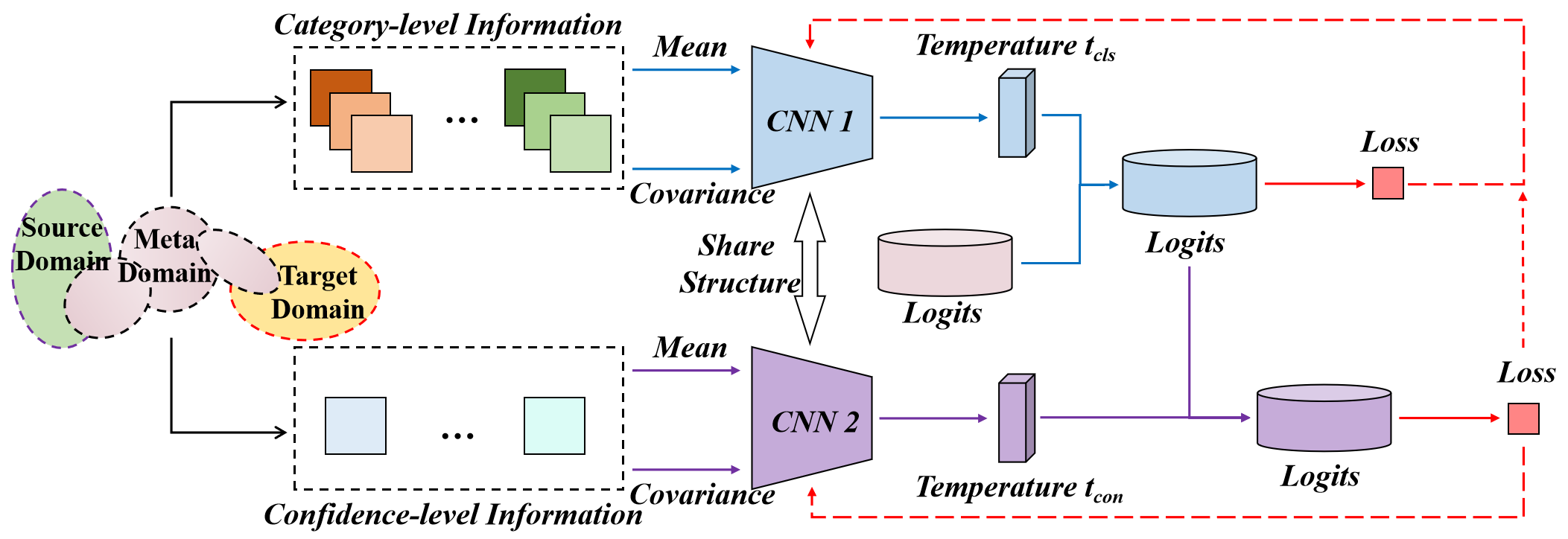}
   \vspace{-0.25in}
   \caption{Overview of the proposed method. Blue/purple lines indicate the pipeline of the category-wise/confidence-level-wise calibration. Red lines indicate the pipeline of the computation to loss and the backpropagation.}
   \label{fig:method}
   \vspace{-0.15in}
\end{figure}

Given a model $\mathcal{F}$ trained from $\mathcal{S}=\{ (x_i,y_i)\}_{i=1}^I$ and a test dataset $\mathcal{D}=\{ (x_n,y_n)\}_{n=1}^N$, $\mathcal{S}$ and $\mathcal{D}$ are both drawn from a joint distribution $\pi (X, Y)$,  where $X \in R^D$ and $Y \in \{1,\ldots, C\}$ are random variables that denote the input space and the label set in the classification task with $C$ classes. 
Denoting the output of $\mathcal{F}$ predicting class $\hat{Y}$ for $X$ as $\mathcal{F}(X)=(\hat{Y},\hat{Z})$, $\hat{Z}$ is transformed into confidence $\hat{P}$ via softmax function $\sigma _{SM}$ as $\hat{P}=\max_c \sigma _{SM}(\hat{Z})^{(c)}$.
Most methods optimize post-hoc calibration functions using a validation set from distribution $\pi (X, Y)$, resulting in low calibration errors for the test data drawn from the same distribution. In this work, we focus on the calibration of prediction uncertainty not only on $\pi (X, Y)$ but also under OOD scenarios with test data from another distribution $\rho(X, Y) \neq \pi (X, Y)$. Fig.~\ref{fig:method} sketches the overall architecture of the proposed method.

\subsection{Subgroup-induced representation} \label{Confidence representation}
Existing methods for OOD post-hoc calibration often necessitate labeled/unlabeled data from the target domain, which might be unavailable in practice~\cite{wang2020transferable,park2020calibrated,pampari2020unsupervised}. Alternatively, perturbing the validation set can calibrate the model without requiring target domain data~\cite{tomani2021post}. Besides, leveraging meta-sets has shown effectiveness in bridging source-target domain gaps~\cite{deng2021does,sun2021label}. 
Motivated by this, we employ meta-sets, formed through diverse geometric/photometric transformations, to mitigate target-source domain distribution differences, enhancing OOD calibration. Denoting meta-sets as $\mathcal{M}=\{\mathcal{M}^1, \cdots, \mathcal{M}^k\}$, where $\mathcal{M}^i$ results from varying transformations on labeled validation sets~\cite{deng2020labels,sun2021label}. These meta-sets train a regression module $\mathcal{G}$, enabling temperature prediction from $\widetilde{D}^T$ drawn from $\rho (X, Y)$.

Due to the high dimensionality, the whole dataset cannot be directly used as the input of $\mathcal{G}$. It is desired to extract a low-dimensional vector as the representation of a dataset. To this end, we design subgroup-induced category-wise and confidence-level-wise representations. 

\textbf{Category-wise representation.} 
We categorize instances by their highest predicted confidence, not ground-truth. Each category ($c \in {1, \cdots, C}$) forms three subgroups ($h_c$, $m_c$, $l_c$) for high, medium, and low confidence. Mean ($\mu(\cdot) \in \mathbb{R}^C$) and covariance ($Cov(\cdot) \in \mathbb{R}^{C\times C}$) are extracted from each subgroup. Combining statistics of all three subgroups, we have the representations $f_{\mu}^c$ and $f_{Cov}^c$ of category $c$: 
\begin{equation}
    f_{\mu}^c=[\mu(l_c),\mu(m_c),\mu(h_c)] \in \mathbb{R}^{3 \times C},
\end{equation}
\begin{equation}
    f_{Cov}^c=[dia(Cov(l_c)),\cdots ,dia(Cov(h_c))] \in \mathbb{R}^{3 \times C},
\end{equation}
where $dia(\cdot)$ takes diagonal elements. $f_{\mu}^c$ signifies subgroup average confidence, and $f_{Cov}^c$ indicates confidence vector variance. Finally, we combine the representations of all categories as the category-wise representation of the dataset:
\begin{equation}
    f_{\mu}=[f_{\mu}^1,\cdots,f_{\mu}^C] \in \mathbb{R}^{C \times 3 \times C},
\end{equation}
\begin{equation}
    f_{Cov}=[f_{Cov}^1,\cdots,f_{Cov}^C ] \in \mathbb{R}^{C \times 3 \times C}.
\end{equation}

\textbf{Confidence-level-wise representation.} 
High confidence scores produce high precision, while most methods assume uniform scaling~\cite{guo2017calibration,jiang2018trust}. Relaxing this constraint is crucial as instances require distinct scaling based on confidence levels. We quantize dataset confidence levels to include more fine-grained information. Here, instances are grouped into $M$ subgroups $B_m, m = 1,\cdots, M$.  
The choice of $M$ is guided by the common practice in the literature, with many related works adopting a division of confidence intervals into 10 segments, then we set $M$=10.
Similar to $f_{\mu}^c$ and $f_{Cov}^c$, mean and covariance vectors are extracted. Thus, we acquire confidence-level-wise representations $z_{\mu}^m=\mu(B_m)\in\mathbb{R}^C$ and $z_{Cov}^m=diag(Cov(B_m))\in\mathbb{R}^{C}$. Combining subgroup representations yields confidence-level-wise representation:
\begin{equation}
    z_{\mu}=[z_{\mu}^1,\cdots,z_{\mu}^M] \in \mathbb{R}^{M \times C},
\end{equation}
\begin{equation}
    z_{Cov}=[z_{Cov}^1,\cdots,z_{Cov}^M] \in \mathbb{R}^{M \times C}.
\end{equation}

\subsection{Cascaded temperature regression mechanism}\label{Secondary calibration mechanism}

\textbf{Category-wise calibration.}
The initial stage of our approach is category-wise calibration. We regress a temperature vector $t_{cls}$ from the category-wise representation, containing temperatures for different category-wise subgroups. In training, only meta-set representations of $\mathcal{M}$ are used, and the scaling is applied to $\mathcal{F}$. The loss function of the regression network $\mathcal{G}_{cls}$ ($CNN1$ in Fig.~\ref{fig:method}) is as follows:
\begin{equation}
    \mathcal{L}_{cls}=\mathcal{L}(\mathcal{F}(x),y,t_{cls}),
\end{equation}
where $t_{cls}=\mathcal{G}_{cls}(f_\mu,f_{Cov}) \in \mathbb{R}^{C}$ and each dimension of $t_{cls}$ is applied to scale the confidence scores of the corresponding predicted category. We use ECE as $\mathcal{L}$ to train $\mathcal{G}_{cls}$.
 
\textbf{Confidence-level-wise calibration.}
Category-wise calibration implies uniform scaling within the same predicted category. However, distinct confidence levels require different scaling. Therefore, we regress temperature vector $t_{con}$ from confidence-level-wise representation to calibrate different confidence-level scores. Similar to $\mathcal{L}_{cls}$, the loss function for $\mathcal{G}_{con}$ ($CNN2$ in Fig.~\ref{fig:method}) is defined as follows:
\begin{equation}
    \mathcal{L}_{con}=\mathcal{L}(\tilde{\mathcal{F}(x)},y,t_{con}),
\end{equation}
where $\tilde{\mathcal{F}(x)}=(\hat{y},\tilde{z})$ denotes the logits $\mathcal{F}(x)=(\hat{y},\hat{z})$ after category-wise calibration. $t_{con}=\mathcal{G}_{con}(z_\mu,z_{Cov}) \in \mathbb{R}^{M}$, and for each confidence level $m \in 1, \cdots, M$, $t_{con}^m$, the $m$-th element of $t_{con}$, is used to perform the corresponding scaling. For instance $x$ in $B_m$, the following scaling performed: 
\begin{equation}
\begin{split}
    \hat{p} 
    &=\max_c \sigma_{SM} \left( \tilde{z}/ t_{con}^{(m)}\right)^{(c)},
\end{split}
\end{equation}
where $\hat{p}$ denotes confidence for instance $x$. Importantly, both category-wise and confidence-level-wise calibrations do not impact overall classification accuracy, as $t_{cls}$ or $t_{con}$ re-scales logits for all classes in each instance.
To jointly train the two regression networks, we define the loss function $\mathcal{L}$ as the sum of $\mathcal{L}_{cls}$ and $\mathcal{L}_{con}$ weighted by $\lambda$ and $(1-\lambda)$, respectively:
\begin{equation}
\mathcal{L}=\lambda \cdot \mathcal{L}_{cls}+(1-\lambda) \cdot \mathcal{L}_{con}.
\label{loss}
\end{equation}

 \section{Experiments}

\begin{table}[t]
\tiny
\setlength{\tabcolsep}{2.5pt}
\begin{center}
\caption{
Comparison of calibration methods. We report ECE (\%)~\cite{naeini2015obtaining} to evaluate the performance, the lower is better.}
\label{results}
\begin{tabular}{l|cccc|ccc|c}
\toprule
Train Set & \multicolumn{4}{c|}{MNIST} & \multicolumn{3}{c|}{CIFAR-10} & TinyImageNet \\ \cline{1-9}
Test Set  & DIGITAL-S & SVHN & USPS & USPS-C & CIFAR-10.1 & CIFAR-10.1-C & CIFAR-F & TinyImageNet-C \\ \hline
Base              & 18.81 & 11.92 & 22.60 & 18.46 & 6.11 & 37.61 & 14.71 & 21.69 \\ 
TS                & 22.20 & 15.51 & 25.54 & 21.57 &  1.69 & 27.85 & 8.57 & 12.41 \\
ETS               & 22.11 & 15.41 & 25.47 & 21.49 &  \textbf{1.37} & 26.26 & 7.44 & 11.95 \\
IRM               & 22.18 & 14.47 & 25.94 & 22.11 &  2.35 & 25.44 & 7.71 & 14.24 \\
IR                & 24.88 & 21.23 & 26.07 & 22.63 &  2.69 & 26.86 & 8.36 & 16.14 \\
TS-IR             & 23.64 & 19.47 & 26.05 & 22.15 &  2.08 & 27.00 & 8.41 & 14.26 \\ \hline
TS-P              & \underline{7.34} & \textbf{0.83} & \underline{9.55} & \underline{7.58} & 8.54 & 16.95 & 3.94 & 6.92 \\
ETS-P             & 9.62 & 2.99 & 12.52 & 9.45 & 8.98 & \underline{16.72} & 4.30 & 6.70 \\
IRM-P             & 9.69 & \underline{2.70} & 13.91 & 10.21 &  6.34 & 17.10 & 3.00 & 6.95 \\
IR-P              & 11.80 & 4.47 & 15.31 & 12.44 &  4.16 & 19.32 & 3.17 & 6.61 \\
TS-IR-P           & 10.76 & 3.95 & 12.77 & 10.81 &  5.93 & 18.03 & \underline{2.86} & \underline{5.49} \\ 
\hline
Ours              & \textbf{4.57} & 4.96 & \textbf{4.54} & \textbf{7.33} &  \underline{1.52} & \textbf{6.28} & \textbf{2.83} & \textbf{3.84} \\ \bottomrule
\end{tabular}
\end{center}
\vspace{-0.15in}
\end{table}

\subsection{Datasets and baselines}
To verify our method, we conduct experiments on image classification tasks (MNIST, CIFAR-10, TinyImageNet). 
Normally, we train a model on a train set, generate meta-sets by transforming the valid set, and evaluate the performance on unseen test sets. 
We compare our approach with the following methods: 
uncalibrated baseline model (Base), TS~\cite{guo2017calibration}, ETS~\cite{zhang2020mix}, IR~\cite{zadrozny2002transforming}, IRM~\cite{zhang2020mix} and TS-IR~\cite{zhang2020mix}. Tune a post-hoc calibrator using the perturbed validation set by a suffix ``-P"~\cite{tomani2021post}.

\subsection{Experimental results}

The overall results are shown in~\tablename~\ref{results}, we report the performance of calibration and have the following observations.

\textbf{Most existing methods are sensitive to OOD scenarios.} 
For instance, TS, ETS, and TS-IR exhibit worse calibration in MNIST, and the performance varies greatly across different test sets. This arises from learning re-scaling functions from the validation set, not accounting for distribution discrepancies. Hence, the above methods falter in OOD scenarios. 

\textbf{Perturbed validation sets can improve calibration under OOD scenarios.}  
For instance, unperturbed baseline results on Digital-S and CIFAR-F exceed 22.11\% and 7.44\%, while perturbed baselines show results below 11.80\% and 4.30\%, respectively. While perturbation enhances learned re-scaling function applicability, performance disparities across test sets persist, such as CIFAR-10.1 and CIFAR-F. This could be attributed to perturbation quality and baseline methods neglecting category and confidence level influence, constraining the expressive power of the learned re-scaling function.

\textbf{Our method is more suitable to calibrate classifier performance under OOD scenarios.} 
In MNIST, we lead on DIGITAL-S, USPS, and USPS-C benchmarks. While SVHN performance slightly lags, the result is still acceptable. 
In CIFAR-10, we dominate CIFAR-10.1-C and CIFAR-F. Notably, our method outperforms others on CIFAR-10.1-C, highlighting its capability for extreme domain shifts. In TinyImageNet, our method reduces ECE from 21.69\% to 3.84\%, far surpassing TS-IR-P at 5.49\%. 
These results support that different categories and different confidence level provide different contributions to the reliability of prediction, which also indicate that the cascaded temperature regression mechanism has a more powerful expressive power to calibrate.

\begin{figure}[!t]
    \centering
    \includegraphics[width=0.99\linewidth]{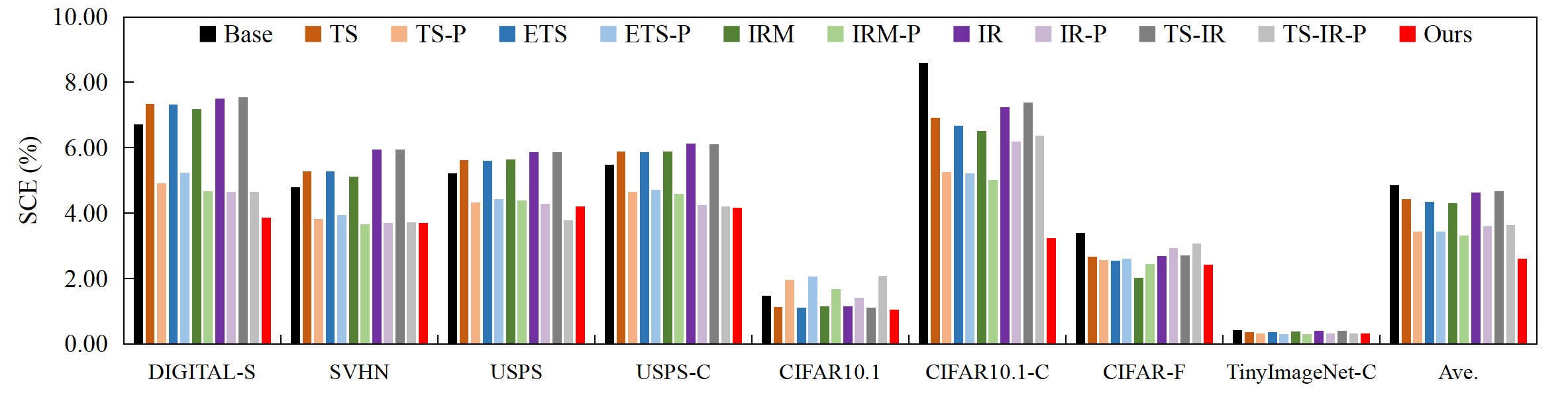}
    \vspace{-0.1in}
    \caption{Comparison of different methods. SCE(\%) is reported. }
    \vspace{-0.1in}
    \label{results_sce}
\end{figure}

\textbf{Result of category-wise calibration performance.}  Static Calibration Error (SCE) is a simple extension of ECE to every probability in the multiclass setting~\cite{nixon2019measuring}. 
Fig.~\ref{results_sce} shows comparative results of SCE under different calibration methods. Compared with the baselines, our approach has largely outperformed baseline post-hoc calibrators, which indicates the category-wise representation can provide fine-grained information describing the category for the calibration task. 

\begin{figure}[!t]
  \centering
    \includegraphics[width=0.24\linewidth]{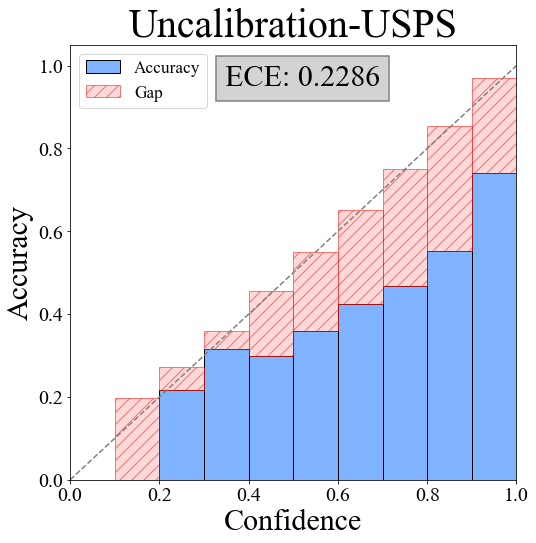}
    \includegraphics[width=0.24\linewidth]{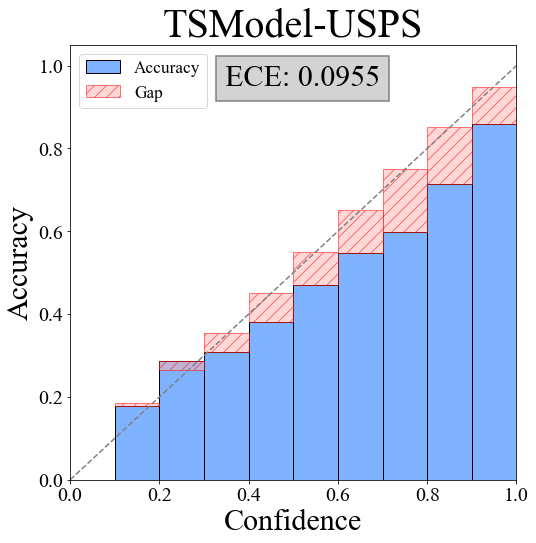}
    \includegraphics[width=0.24\linewidth]{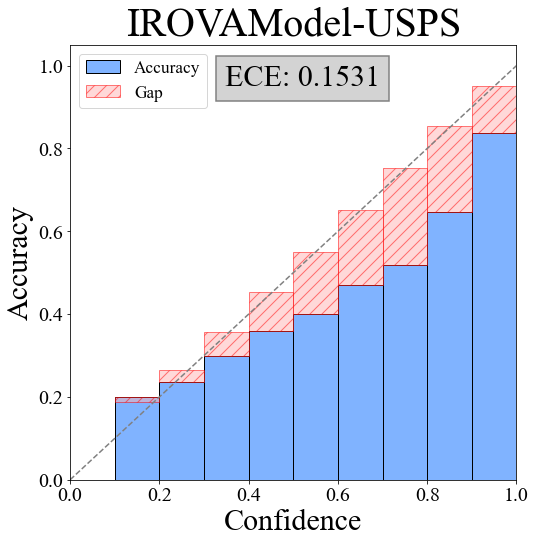}
    \includegraphics[width=0.24\linewidth]{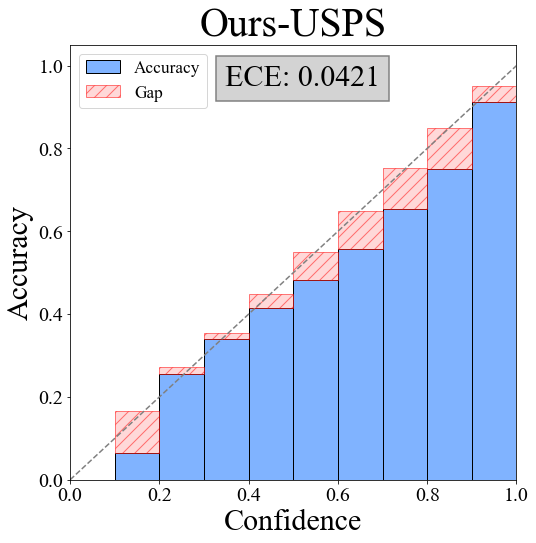}
    \vspace{-0.2in}
   \caption{
   Reliability diagram on USPS using different methods. Baselines are trained on the perturbed validation set.
   }
   \vspace{-0.15in}
   \label{fig:result_compare}
\end{figure}

\textbf{Result of confidence-level-wise calibration performance.} 
We illustrate the calibration performance of different methods on USPS in Fig.~\ref{fig:result_compare}. As expected, compared to baselines, our method improves the calibration error for post-hoc calibrators and the result is consistent with the notion that different confidence level requires different optimal scaling.

\begin{table}[!t]
\tiny
\setlength{\tabcolsep}{3pt}
\begin{center}
\caption{
Ablation results for our methods. ECE(\%) is shown. 
}
\label{ablation}
\begin{tabular}{l|cccc|ccc}
\toprule
Train Set & \multicolumn{4}{c|}{MNIST} & \multicolumn{3}{c}{CIFAR-10} \\ \cline{1-8}
Test Set   & DIGITAL-S & SVHN & USPS & USPS-C & CIFAR-10.1 & CIFAR-10.1-C & CIFAR-F\\ \hline
$t_{cls}$ W/o $f_{\mu}$ &\underline{3.99} &5.80 &6.76 &7.07 &1.88 &\underline{5.06} &3.44 \\ 
$t_{cls}$ W/o $f_{Cov}$ &5.65 &5.80 &\underline{4.87} &8.51 &\underline{1.13} & 7.25 & 3.98 \\ 
W/o $t_{cls}$           &5.80 &5.04  &6.91 &8.06 &1.50 &7.28 &\underline{2.60}\\ \hline
$t_{con}$ W/o $z_{\mu}$ &5.15 &\underline{4.15} &4.50 &\underline{7.02} &1.55 &6.73 &4.54\\ 
$t_{con}$ W/o $z_{Cov}$ &5.03 &5.64 &5.98 &7.29 &2.06 &9.04 &4.91\\ 
W/o $t_{con}$           & 4.59 &4.39 &7.27 &8.42 &1.92 &7.34 &3.90\\ \hline
Ours & \textbf{3.86} & \textbf{3.69} & \textbf{4.21} & \textbf{4.17} & \textbf{1.04} & \textbf{3.24} & \textbf{2.43} \\ \bottomrule
\end{tabular}
\end{center}
\vspace{-0.2in}
\end{table}

\subsection{Ablation study}
\textbf{Category-wise and confidence-level-wise representation.} 
\tablename~\ref{ablation} shows that distinct components offer varying dataset descriptions, with optimal outcomes when both representations are used simultaneously. This indicates both representations provide diverse fine-grained information for calibration, supporting that distinct categories or confidence levels contribute differently to the reliability of the model’s prediction.

\begin{figure}[t]
  \centering
      \includegraphics[width=0.34\linewidth]{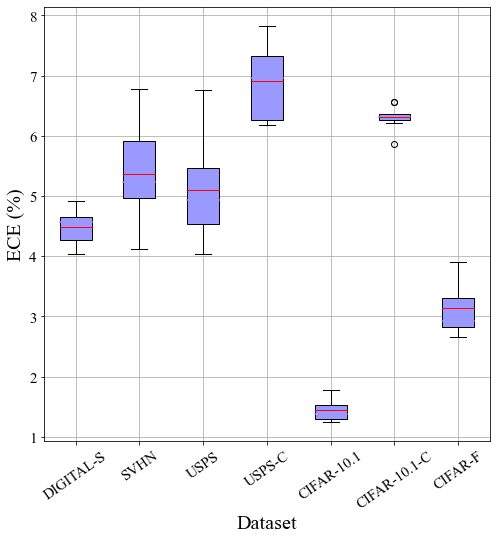}
       \includegraphics[width=0.63\linewidth]{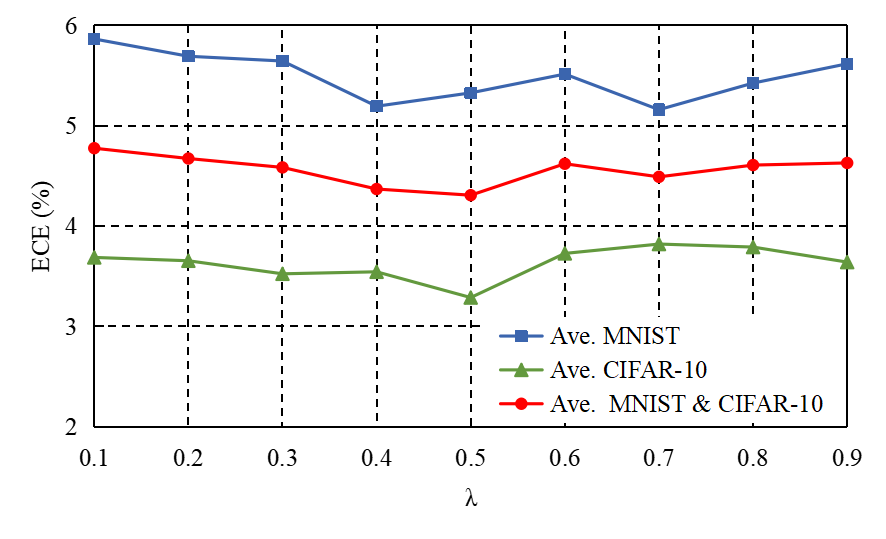}
  \vspace{-0.1in}
  \caption{Impact of the fusion coefficient $\lambda$. Left: Variation of ECE for each dataset with different $\lambda$. Right: The average ECE of the group's test datasets of MNIST and CIFAR-10.}
  \vspace{-0.15in}
   \label{fig:ablation}
\end{figure}

\textbf{The influence of hyper-paramter $\lambda$.} 
$\lambda$ balances the influence between the category-wise information and the confidence-level-wise information. 
In \figurename~\ref{fig:ablation}, ECE variations across different $\lambda$ values are depicted during the MNIST and CIFAR-10 experiment setup. 
Largest individual set fluctuation is 2.72\% (USPS), while group sets fluctuate is up to 0.71\% (Ave. MNIST). 
Therefore, our method is resilient to the variations of $\lambda$, and we use $\lambda=0.4$ for the experiment.
 \section{Conclusion}
In this paper, we propose a novel cascaded temperature regression method to realize OOD calibration. Different from previous works, our method incorporates the confidence and predicted category information to form subgroup-wise representations and regresses different temperatures for different subgroups in diverse domains. Extensive experiments across three benchmarks show the effectiveness and superiority of our method compared to various alternative approaches.

\begin{sloppypar}
\noindent\textbf{ACKNOWLEDGEMENT.} This work was supported in part by the National Natural Science Foundation of China No. 62376277, No. 61976206 and No. 61832017, Beijing Outstanding Young Scientist Program NO. BJJWZYJH012019100020098, the Fundamental Research Funds for the Central Universities, the Research Funds of Renmin University of China 21XNLG05.
\end{sloppypar}

\bibliographystyle{IEEEbib}
\bibliography{strings,refs}

\end{document}